\documentclass[10pt,twocolumn,letterpaper]{article}

\usepackage{iccv}              

\newcommand{\reffig}[1]{Figure~\ref{fig:#1}}

\newcommand{\reftbl}[1]{Table~\ref{tab:#1}}

\newcommand{\ignorethis}[1]{}

\makeatletter
\DeclareRobustCommand\onedot{\futurelet\@let@token\@onedot}
\def\@onedot{\ifx\@let@token.\else.\null\fi\xspace}

\makeatother

\usepackage[accsupp]{axessibility}  

\definecolor{iccvblue}{rgb}{0.21,0.49,0.74}
\usepackage[pagebackref,breaklinks,colorlinks,allcolors=iccvblue]{hyperref}
\usepackage{pifont} 
\usepackage{makecell}

\def\paperID{12919} 
\def\confName{ICCV}
\def\confYear{2025}

\title{Integrating Visual Interpretation and Linguistic Reasoning \\ for Math Problem Solving}

\author{
  {\small Zixian Guo\textsuperscript{\rm 1,2} \quad Ming Liu\textsuperscript{\rm 1} \quad Qilong Wang\textsuperscript{\rm 3} \quad Zhilong Ji\textsuperscript{\rm 4} \quad Jinfeng Bai\textsuperscript{\rm 4} \quad Lei Zhang\textsuperscript{\rm 2}\quad  Wangmeng Zuo\textsuperscript{\rm 1,5} } \\ 
  {\small \textsuperscript{\rm 1}Harbin Institute of Technology \quad \textsuperscript{\rm 2} The Hong Kong Polytechnic University \quad \textsuperscript{\rm 3}Tianjin University} \\ 
  {\small \textsuperscript{\rm 4}Tomorrow Advancing Life \quad \textsuperscript{\rm 5}Pazhou Lab, Guangzhou}
}

\begin{document}
\maketitle
\begin{abstract}

\vspace{-1em}

Current large vision-language models (LVLMs) typically employ a connector module to link visual features with text embeddings of large language models (LLMs) and use end-to-end training to achieve multi-modal understanding in a unified process. Effective alignment needs high-quality pre-training data and a carefully designed training process.
Current LVLMs face challenges when addressing complex vision-language reasoning tasks, with their reasoning capabilities notably lagging behind those of LLMs.
This paper proposes a paradigm shift: instead of training end-to-end vision-language reasoning models, we advocate for developing a decoupled reasoning framework based on existing visual interpretation specialists and text-based reasoning LLMs.
Our approach leverages (1) a dedicated vision-language model to transform the visual content of images into textual descriptions and (2) an LLM to perform reasoning according to the visual-derived text and the original question.
This method presents a cost-efficient solution for multi-modal model development by optimizing existing models to work collaboratively, avoiding end-to-end development of
vision-language models from scratch. By transforming images into language model-compatible text representations, it facilitates future low-cost and flexible upgrades to upcoming powerful LLMs.
We introduce an outcome-rewarded joint-tuning strategy to optimize the cooperation between the visual interpretation and linguistic reasoning model.
Evaluation results on vision-language benchmarks demonstrate that the decoupled reasoning framework outperforms recent LVLMs. 
Our approach yields particularly significant performance gains on visually intensive geometric mathematics problems. 
The code is available: https://github.com/guozix/DVLR.

\end{abstract}
\begin{figure*}[t!]
  \centering
  \includegraphics[width=1\linewidth]{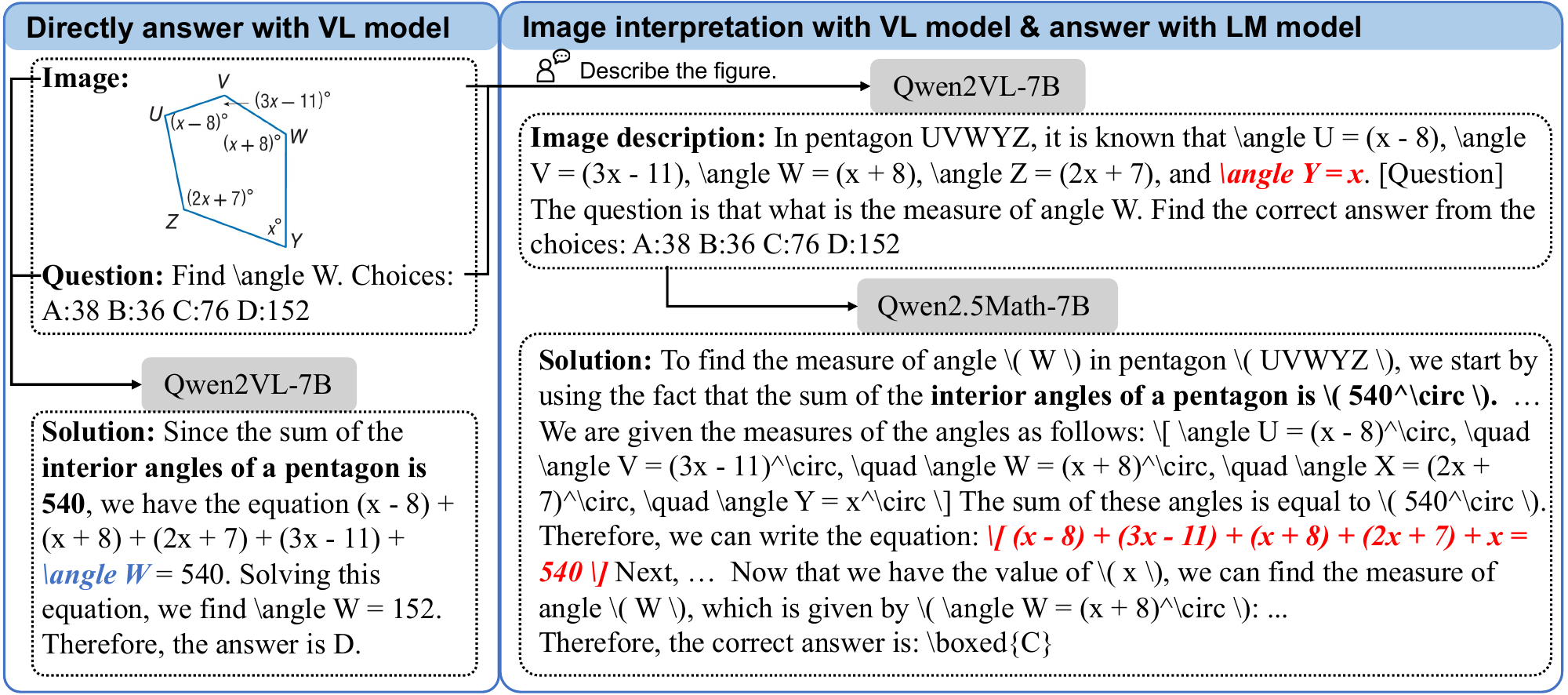}
  \vspace{-1em}
  \caption{Comparison of the solving results of a geometry math problem by directly answering with a VL model or first interpreting the image with a VL model and then answering with an LM model. We can see that the VL model can correctly describe the image even though it can not correctly answer the question. Decoupled reasoning architecture with an image interpreter and a linguistic reasoner can solve the problem correctly.
}
  \label{fig:intro}
  \vspace{-1em}
\end{figure*}

\vspace{-1em}
\section{Introduction}
\label{sec:intro}
Recent application of reinforcement techniques to the learning of large language models (LLMs) triggers amazing reasoning ability~\cite{luongReFTReasoningReinforced2024, rafailov2023direct} in the form of long natural language thought or slow-thinking~\cite{xiangAtomThinkSlowThinking2024}.
Much efforts have been made to extend the powerful reasoning capability to visual and textual multi-modal scenarios~\cite{haoCanMLLMsReason2025, chengCoMTNovelBenchmark2024,gaoInterleavedModalChainofThought2024,zhangEuclidSuperchargingMultimodal2024,zhangHowFarAre2024}. Current development of vision-language models typically employ a connector module, e.g. Linear layers~\cite{llava}, Q-Former~\cite{blip2}, Perceiver-Resampler~\cite{alayrac2022flamingo}, etc., to integrate visual features with text features of LLMs. They collect high-quality image-text paired data and design sophisticated end-to-end training processes to achieve joint visual and textual comprehension and reasoning in a unified process~\cite{xuLLaVAo1LetVision2024, yaoMulberryEmpoweringMLLM2024}.

However, compared to LLMs, existing large vision-language models (LVLMs) can hardly achieve the same level of reasoning proficiency\cite{zhang2024improve, wuRoleChainofThoughtComplex2023}. This gap in performance is particularly evident in visual-intensive tasks that require precise extraction and interpretation of visual information to support complex reasoning processes~\cite{huangWhyVisionLanguage2025}. 
We find that a main barrier that hinders effective reasoning is that existing LVLMs struggle with precisely extracting the necessary information from images. The inaccurate understanding of the image brings a bottleneck to the subsequent reasoning process, such as hallucinations in the reasoning.

For example, in \reffig{intro}, the foundation model Qwen2VL-7B~\cite{qwen2vl} can not correctly answer the question due to incorrectly associating the variable $x$ with the angle $W$. However, if we prompt the VL model to describe the figure of the question, it can precisely summarize all the interior angles correctly. Then, we can get the correct answer by using a reasoning LM to derive the solution. We show more similar cases in the supplementary material. Even when the images are frustratingly simple, LVLM can still make obvious mistakes if it answers the question directly. This implies that some parts of the current problems in VL reasoning are not actually reasoning problems but insufficient visual understanding.

This observed discrepancy motivates our thinking of whether an end-to-end visual-language reasoning model is optimal or whether the problem can be more effectively addressed through a division of labor between vision specialists and linguistic reasoners. Current LLMs have reached unprecedented capabilities in textual reasoning~\cite{shao2024deepseekmath,guo2025deepseek}, while specialized vision models demonstrate remarkable capabilities in fine-grained image parsing. Rather than pursuing unified training of multi-modal reasoning systems from scratch, it is promising to leverage the complementary strengths of these mature, specialized models to advance visual-language reasoning.


To this end, we propose a decoupled vision-language reasoning architecture that explicitly decouples and orchestrates visual interpretation and linguistic reasoning. 
Unlike common supervised fine-tuning pipelines, which often train LVLMs to reason and solve problems directly with vision-language input, our method introduces an intermediary step where the LVLM first translates visual content into reasoning context. This translation involves extracting key elements, such as geometric primitives and spatial relationships, into natural language descriptions for the subsequent reasoning phase.
The linguistic reasoning is then carried out by an LLM, which processes the natural language conditions aggregated from the input text and image contents to derive the solution through step-by-step reasoning. 
This specialized division of labor between visual interpretation and linguistic reasoning effectively leverages the complementary strengths of each component.

To achieve an effective collaboration of visual interpretation and linguistic reasoning, we built a three-stage training process for LLMs and LVLMs, with each stage addressing different aspects. In the first stage, we separately tune the LVLMs and LLMs to develop the basic ability to parse images and answer questions, respectively. Based on existing geometric math problem datasets, we built a dataset of geometric diagrams and interpretation text pairs to train the LVLM to parse images. And we used the solution steps to train the LLM to answer the question. We then connect the LVLM and LLM in series. In the next training stage, we use the correctness of the answer by the LLM as reward feedback to continue optimizing the LVLM, further enabling it to generate more effective image parsing results for LLM reasoning. In the third stage, we fix the LVLM and further tune the LLM to enhance the reasoning performance by rewarding the successful outcome of the response as the reinforced learning objective. This joint tuning strategy ensures both modules collaborate effectively.

Our approach yields particularly significant performance gains on visually intensive geometric mathematics problems. 
In summary, our contributions include:

\begin{itemize}
    \item We propose a novel decoupled reasoning framework for vision-language reasoning that integrates the strengths of LVLM and LLM for precise image interpretation and linguistic reasoning. We suggest that utilizing the ability of mature specialized models forms a valuable baseline for vision-language reasoning.
    \item We propose an effective outcome-rewarded joint optimization approach that facilitates the seamless collaboration between the visual interpretation and linguistic reasoning modules.
    \item Our method demonstrates improvements in vision-language reasoning tasks, particularly visually intensive geometric math problems. This advancement highlights the effectiveness of the decoupled reasoning architecture.
\end{itemize}

\begin{figure*}[t!]
  \centering
  \includegraphics[width=0.8\linewidth]{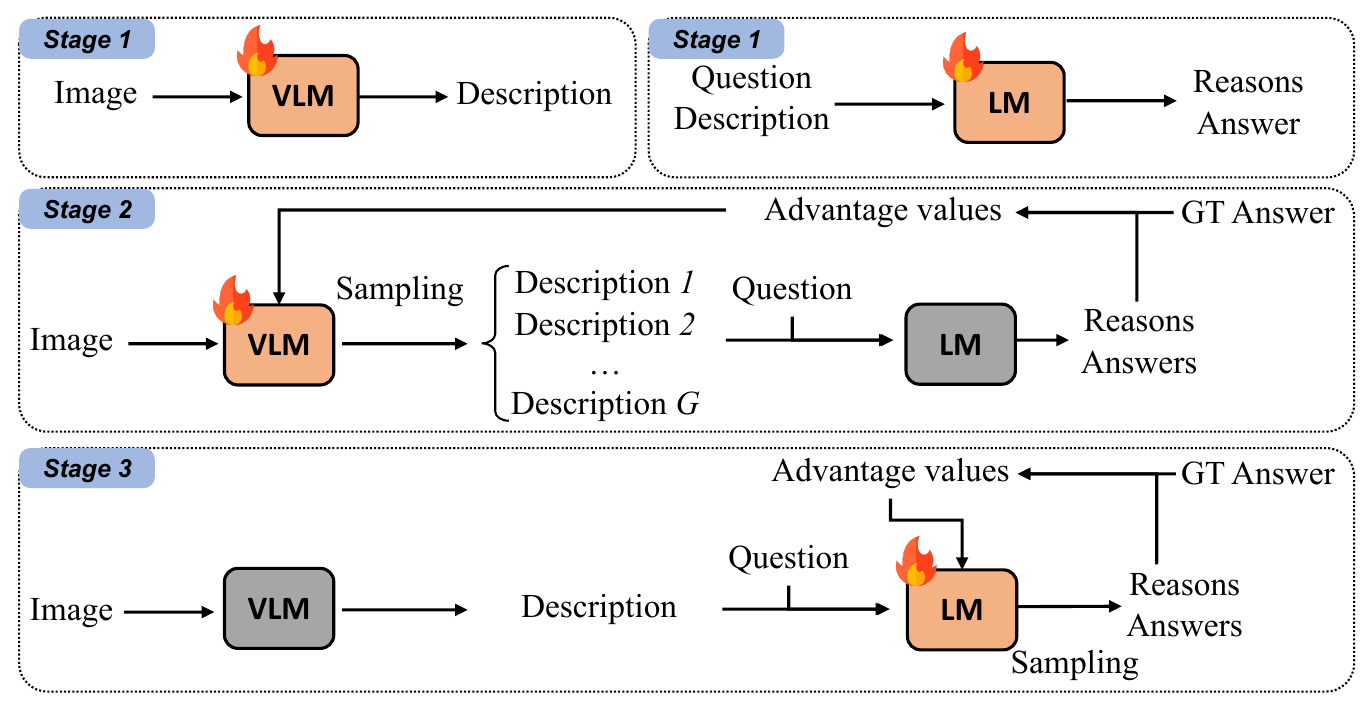}
  \vspace{-1em}
  \caption{The three-stage model training pipeline of our method. The first supervised fine-tuning stage establishes a foundation of both the LVLM and the LLM. The reinforced training stage two and stage three jointly tune the LVLM and LLM to collaborate to solve the geometry problem.
}
  \label{fig:method}
  \vspace{-1em}
\end{figure*}

\section{Related Work}

\subsection{Large Vision-Language Model}
The integration of language and visual content has garnered significant attention in recent years, leading to the rapid advancement of large vision-language models (LVLMs). 
These models aim to bridge the gap between visual and textual modalities, facilitating a more nuanced understanding of multimodal data. Notable methodologies include the Q-former proposed by the BLIP series~\cite{blip,blip2,instructblip}, which utilizes a query-based approach to align visual features with language tokens. Similarly, the LLaVA series~\cite{llava,llava1.5} and MiniGPT4~\cite{zhu2023minigpt} leverage the linear layer to enhance the transition between visual and language tokens. The perceiver resampler used by Flamingo~\cite{alayrac2022flamingo} also shows effective handling of visual content. 

Despite these advancements, achieving a fine-grained joint understanding of visual and linguistic information remains a difficult challenge. Recent models, such as the Qwen-VL series~\cite{qwenvl,qwen2vl}, have introduced innovative mechanisms like dynamic visual tokens that adapt to varying image resolutions. There is also work exploring more effective image feature encoding mechanisms~\cite{weiSlowPerceptionLets2024}, which gradually extract more comprehensive visual features from coarse to fine granularity. These enhancements are crucial for improving the accuracy of visual perception and mitigating issues associated with hallucinations, which have historically advanced LVLMs.

\subsection{Vision-Language Reasoning}
Recently, reinforcement-learning-based fine-tuning~\cite{luongReFTReasoningReinforced2024,lyuExploringLimitOutcome2025,hosseiniVSTaRTrainingVerifiers2024, kumarTrainingLanguageModels2024}, test-time scaling~\cite{muennighoffS1SimpleTesttime2025, snellScalingLLMTestTime2024a} and learning from tree search~\cite{yaoMulberryEmpoweringMLLM2024, zhangLLaMABerryPairwiseOptimization2024,qiMutualReasoningMakes2024} techniques have enabled impressive progress on complex reasoning tasks with large language models~\cite{yu2023metamath, wangMathShepherdVerifyReinforce2024, shao2024deepseekmath}. This has inspired people to have a great desire for multimodal joint reasoning of vision and language. However, the interplay between visual perception and reasoning often leads to instability. Fortunately, the increasing availability of high-quality visual-language reasoning datasets~\cite{gaoGLLaVASolvingGeometric2023, dengRCoTReverseChainThought2024} has significantly improved the reasoning capabilities of LVLMs.

Geometric math problems, characterized by their visual and linguistic components, have emerged as a focal point for research, as they allow for straightforward verification of results. Various approaches have been proposed, including G-LLaVA~\cite{gaoGLLaVASolvingGeometric2023}, Math-LLaVA~\cite{shiMathLLaVABootstrappingMathematical2024}, MultiMath~\cite{pengMultiMathBridgingVisual2024}, LLaVA-o1~\cite{xuLLaVAo1LetVision2024}, which utilize large language models to expand the diversity of existing datasets by rephrasing and rewriting problems and reasoning steps. InfiMM-WebMath~\cite{hanInfiMMWebMath40BAdvancingMultimodal2024} collected a large number of documents containing mathematical knowledge from the Internet. R-COT~\cite{dengRCoTReverseChainThought2024} proposed to synthesize high-quality math reasoning problems by reversely constructing question-answering pairs by prompting LLM with the description of the geometry image. This methodology underscores the effectiveness of supervised fine-tuning with data derived from diverse sources to create capable LVLMs adept at solving graphic mathematical challenges.

Recent efforts to enhance reasoning in large multimodal language models have focused on learning from tree-based exploration~\cite{qinO1ReplicationJourney2024,yaoMulberryEmpoweringMLLM2024} and reinforcement learning~\cite{lightmanLetsVerifyStep2023a}.
For instance, Mulberry~\cite{yaoMulberryEmpoweringMLLM2024} proposed a collective-learning-based MCTS framework to explore feasible reasoning paths efficiently and effectively. Learning from the set of well-defined reasoning steps and reflection paths achieves multimodal LLMs with o1-like step-by-step reasoning and reflection capabilities. 

As the field continues to evolve, the integration of more effective training methodologies and the development of high-quality datasets will be paramount in advancing the capabilities of LVLMs. Our proposed decoupled reasoning architecture builds upon these foundational works by explicitly decoupling visual interpretation from linguistic reasoning, thereby addressing limitations observed in existing models and paving the way for enhanced performance in complex geometric problem-solving tasks.

\section{Method}
\vspace{-0.5em}
\subsection{Overview}
\vspace{-0.5em}
In this section, we introduce how we build the decoupled reasoning vision-language reasoning framework. The training process is divided into three stages. The first stage establishes a robust foundation by supervised fine-tuning of both the LVLM and the LLM. This stage ensures that each module develops the necessary capabilities to perform its designated role effectively. The second stage further reinforces the LVLM to generate informative descriptions of images for the LLM to derive accurate answers. The third stage aims to refine the LLM's ability to reason about the visual information provided by the LVLM and derive accurate solutions through step-by-step reasoning. The overall training pipeline is shown in \reffig{method}.

\vspace{-0.5em}
\subsection{Supervised Fine-tuning of LVLM and LLM}
\vspace{-0.5em}

\textbf{Image Interpretation Dataset.}
Existing vision-language problem solving datasets~\cite{dengRCoTReverseChainThought2024, gaoGLLaVASolvingGeometric2023, shiMathLLaVABootstrappingMathematical2024, pengMultiMathBridgingVisual2024} are usually in the form of problems and solutions. Problems include images, corresponding questions, and solutions include reasoning steps and final answers.
To train the LVLM on the intermediary image interpretation task, we construct a specialized dataset sourced from GeoMM~\cite{dengRCoTReverseChainThought2024}. GeoMM generates geometry diagrams by sequential sampling and combining basic shapes. At the same time, it also produces a description of the connection and relation of the shapes as it generates the geometric figure. This description data is used to synthesize questions and answers to fine-tune LVLMs. We believe that the accurate descriptions paired with the images are of great value in training the LVLM to comprehensively understand images.

\begin{figure*}[t!]
  \centering
  \includegraphics[width=0.9\linewidth]{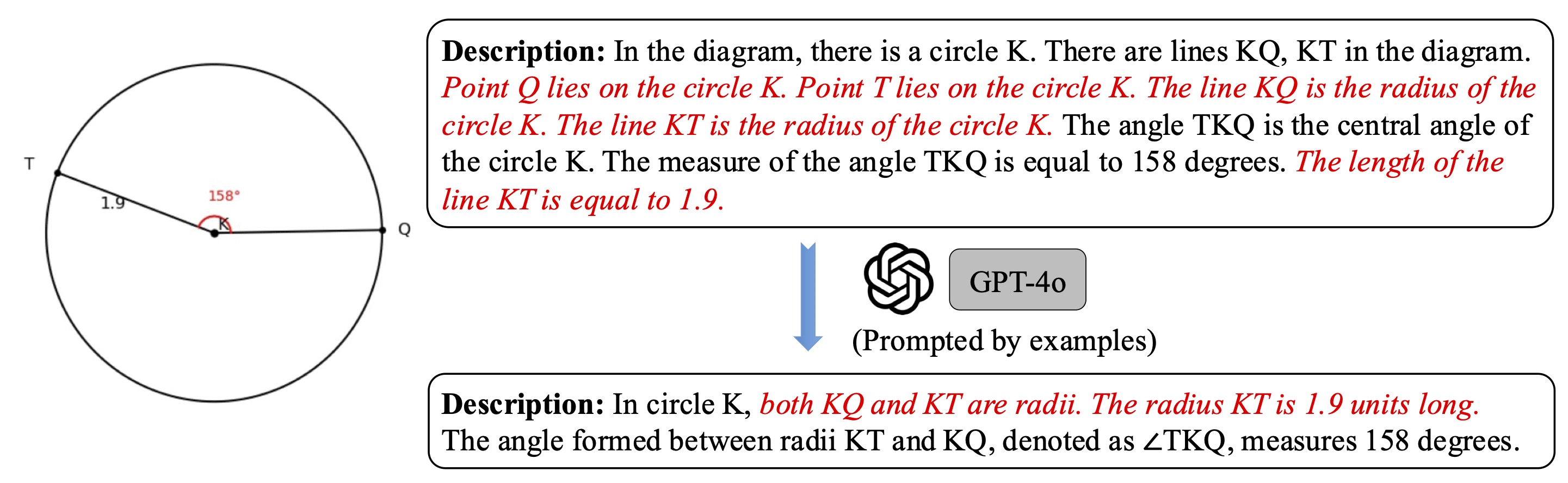}
  \vspace{-1em}
  \caption{Demonstration of the geometry image description data. The original image description from the GeoMM dataset is accurate but wordy and lacks logical order. We refined the description to be more concise and logical by prompting GPT-4o.
}
  \label{fig:data}
  \vspace{-1em}
\end{figure*}

Although these natural language descriptions contain accurate geometric primitives (e.g., points, lines, angles) and detailed spatial relationships (e.g., parallelism, perpendicularity) present in the diagrams, they are not optimal since they are generated according to rules. These descriptions tend to be long-winded and lack logical order, which leads to suboptimal training results, as shown in \reffig{data}.To solve this problem, we preprocess the descriptions by prompting GPT-4o to refine the descriptions by removing duplicate contents and improving description order while retaining all relevant visual information.

In addition, in order to deal with the situation where some questions in the visually intensive benchmark test samples are printed in the image, we also augment the images in the dataset. We embed text questions into geometric images and train the model to accurately read the questions and parse the geometric structures in the images. Thanks to the accuracy of the original description and the effective data construction method, we have obtained a high-quality geometric image content parsing dataset.

\noindent \textbf{Learning to Interpret Images.}
Using the constructed image content interpretation data, we supervise fine-tune the LVLM to conduct two related tasks. The first is to parse the question text embedded in the image. After reading the question, the second task is to generate a description of the geometric shapes in the image. During testing, the first task is omitted for those testing samples that have provided the question text explicitly. This supervised fine-tuning process equips the LVLM with the basic ability to read the question and identify useful information for solving the question from the image.
\noindent \textbf{Learning to Reasoning with Visual-Derived Texts.}
In parallel, we supervise fine-tune the LLM to generate step-by-step solutions depending on the natural language formalized visual conditions. We use the constructed image descriptions and target questions as input and the corresponding solution steps that originate from GeoMM dataset as output to train the LLM, ensuring that the LLM can respond with the desired format (step-by-step reasoning and answer in the end) and logical reasoning steps.

By separating the tasks of visual interpretation and linguistic reasoning, we enable each module to specialize in its respective domain, laying the groundwork for effective collaboration in the subsequent stages of training.

\vspace{-0.5em}
\subsection{Reinforced Interpretation Learning of LVLM}
\vspace{-0.5em}

After the first training stage, we connect the LVLM and LLM in series. Then, we proceed to the second stage of training, which focuses on reinforcing the LVLM's ability to generate more informative descriptions of geometric images. This stage employs a reinforcement learning (RL) algorithm, where the correctness of the LLM's final answer serves as the reward signal for optimizing the LVLM.

Specifically, the LVLM is regarded as the policy model $\pi_{\theta}$. We optimize the policy model by maximizing the objective given by the Group Relative Policy Optimization (GRPO) algorithm~\cite{shao2024deepseekmath}:

\vspace{-0.5em}
\begin{equation}
\begin{split}
\mathcal{J}_{G R}& _{P O} (\theta) =\mathbb{E}\left[q \sim \mathcal{D},\left\{o_{i}\right\}_{i=1}^{G} \sim \pi_{\theta_{o l d}}(O \mid q)\right] \cdot \\
& \frac{1}{G} \sum_{i=1}^{G} (\min (\frac{\pi_{\theta}\left(o_{i} \mid q\right)}{\pi_{\theta_{o l d}}\left(o_{i} \mid q\right)} A_{i}, \operatorname{clip}(\frac{\pi_{\theta}\left(o_{i} \mid q\right)}{\pi_{\theta_{o l d}}\left(o_{i} \mid q\right)}, \\
& 1-\varepsilon, 1+\varepsilon) A_{i} )-\beta \mathbb{D}_{K L}\left(\pi_{\theta} \| \pi_{r e f}\right) )
\end{split}
\end{equation}
\vspace{-0.5em}

\noindent where the $q$ denotes the question from training dataset $\mathcal{D}$, $\left\{o_{i}\right\}_{i=1}^{G}$ are sampled image description outputs from the LVLM, $\varepsilon$ and $\beta$ are hyper-parameters, and $\left\{A_{i}\right\}_{i=1}^{G}$ is the advantage values computed by normalizing a group of reward values $\left\{r_{i}\right\}_{i=1}^{G}$:

\vspace{-1em}
\begin{equation}
\begin{split}
A_i = \frac{r_i - \operatorname{mean}\left( r_1, r_2, \cdots, r_G \right)}{\operatorname{std}\left( r_1, r_2, \cdots, r_G \right)}
\end{split}
\end{equation}
\vspace{-1em}

\noindent The $KL$ regular term is formulated as :

\vspace{-1em}
\begin{equation}
\mathbb{D}_{K L}\left(\pi_{\theta}| | \pi_{r e f}\right)=\frac{\pi_{r e f}\left(o_{i} \mid q\right)}{\pi_{\theta}\left(o_{i} \mid q\right)}-\log \frac{\pi_{r e f}\left(o_{i} \mid q\right)}{\pi_{\theta}\left(o_{i} \mid q\right)}-1,
\end{equation}
\vspace{-1em}

In this setup, the LVLM generates several descriptions for the geometric diagram of a training sample. Then, these descriptions, coupled with the text question, are passed to the frozen LLM to solve the problem. The LLM's outputs, which contain reasoning steps and the final answer in the end, are evaluated against the ground truth answer. In this training stage, we use multiple-choice questions from the Geo170k dataset~\cite{gaoGLLaVASolvingGeometric2023} so that the expected answer is one of the options provided in the question. We extract the final answer from the generated response by regular expressions and then directly compare the generated answer with the ground truth. A rule-based reward function is used according to the existing method~\cite{luongReFTReasoningReinforced2024}. The reward $r_i$ of the image description $o_i$ is $1$ if the comparison of the answer is successful. Otherwise, the reward is set as 0.

To make the optimization effective, we skip the training questions where LM is successful (or unsuccessful) on all the sampled descriptions from LVLM. In other words, if $\left\{r_{i}\right\}_{i=1}^{G}$ are all equal to $1$ or $0$, we skip the sample. This reward value design strategy gives feedback to the LVLM, encouraging it to produce descriptions that are more conducive to accurate reasoning by the LLM. This outcome-rewarded tuning strategy ensures that the LVLM learns to prioritize the extraction of visual information that is most relevant for solving the problem.

\begin{figure}[t!]
  \centering
  \includegraphics[width=0.98\linewidth]{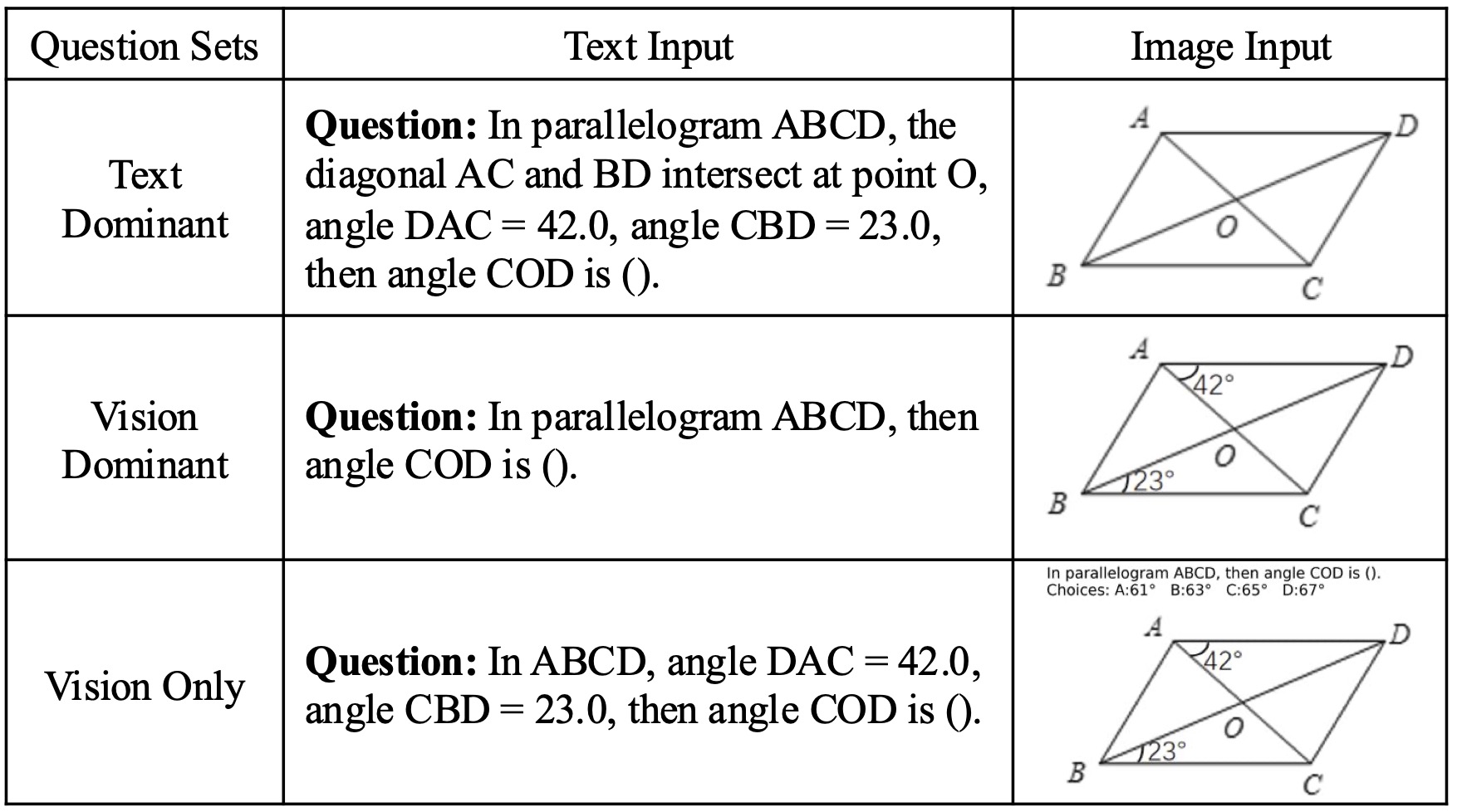}
  \vspace{-1em}
  \caption{Visualization of different versions of the same problem in MathVerse testmini.
}
  \label{fig:verse}
  \vspace{-1em}
\end{figure}

\begin{table*}[t]
    \centering
    
    \caption{Comparison of unified vision-language reasoning model LLaVA-CoT and the decoupled reasoning framework.}
    \vspace{-1em}
    \label{tab:pre_res}
    \resizebox{0.95\linewidth}{!}{
    \setlength{\tabcolsep}{7pt}
    \begin{tabular}{lcccc}
        \toprule
        {Model} & MMStar & MM-Vet & MathVista & MathVerse \\ \hline
        LLaVA-CoT-11B~\cite{xuLLaVAo1LetVision2024} & 57.6 & 60.3 & 54.8 & 36.2 \\  
        Ours-SFT (Init. by Qwen2VL-7B \& Qwen2.5Math-7B) & 59.1 &  61.5 & {57.1} & {42.5} \\

        \bottomrule
    \end{tabular}
    }
    \vspace{-1em}
\end{table*}

\begin{table*}[t]
    \centering
    \caption{Top-1 Accuracy (\%) on geometry math problem solving on the testmini set of MathVerse benchmark.}
    \vspace{-1em}
    \label{tab:main_res}
    \setlength{\tabcolsep}{7pt}
    \begin{tabular}{lccccc}
        \toprule
        {Model} & {\makecell[c]{Text \\ Dominant}} & {\makecell[c]{Text \\ Lite}} & \makecell[c]{Vision\\ Intensive} & \makecell[c]{Vision\\ Dominant} & \makecell[c]{Vision \\ Only} \\ \hline
        \multicolumn{6}{c}{\textit{Closed-source Models}}\\ 
        \hline
        GPT-4o & 54.8 & 52.3 & 47.3 & 44.3 & 43.7 \\  
        
        
        GPT-4o \& OpenAI-o1mini & \textbf{70.6} & \textbf{67.3} & \textbf{64.2} & \textbf{62.0} & \textbf{60.1} \\ 
        
        \hline
        \multicolumn{6}{c}{\textit{Open-source Models}}\\ \hline

        G-LLaVA-7B~\cite{gaoGLLaVASolvingGeometric2023} & 24.9 & 22.1 & 18.0 & 15.2 & 9.0 \\
        Math-LLaVA-13B~\cite{shiMathLLaVABootstrappingMathematical2024} & 22.8 & 21.8 & 21.1 & 19.2 & 15.4 \\
        MultiMath-7B~\cite{pengMultiMathBridgingVisual2024} & 34.8 & 30.8 & 28.1 & 25.9 & 15.0 \\
        R-CoT-7B~\cite{dengRCoTReverseChainThought2024} & 52.3 & 48.2 & 42.6 & 38.5 & 28.5 \\ 
        
        Qwen2VL-7B & 44.3 & 41.5 & 38.2 & 37.1 & 33.8 \\  
        
        
        Qwen2VL-72B \& Deepseek-V3(671B) & \textbf{59.1} & \textbf{53.1} & \textbf{48.9} & 44.2 & 42.9 \\ 

        \hline 
        
        Ours (Init. by Qwen2VL-7B \& Qwen2.5Math-7B) & 54.3 & 49.0 & 46.3 & \textbf{47.2} & \textbf{43.8} \\

        \bottomrule
    \end{tabular}
    \vspace{-1em}
\end{table*}

\subsection{Reinforced Reasoning of LLM}
In the final stage of training, we aim to stimulate the ability of linguistic reasoning further. We fix the parameters of the LVLM and focus on optimizing the LLM's reasoning capabilities. We use the same reward-model-free RL framework as the second training stage. However, in this training stage, the outcome-based reward is given to the sampled responses of the LLM policy model. This training stage refines the LLM's ability to process image interpretations and derive accurate solutions. 
For the questions that can not be correctly responded to even once in the second train stage, the exploration in the LLM sampling process can discover reasoning paths that lead to correct answers.

Through the joint outcome-rewarded tuning process in stage 2 and stage 3, the LVLM and the LLM can collaborate more effectively, enhancing the overall performance of the decoupled reasoning architecture.

\section{Experiment}
\vspace{-0.5em}
\subsection{Setup}
\vspace{-0.5em}
\textbf{Base model.}
We conduct experiments with Qwen2VL-7B~\cite{qwen2vl} and Qwen2.5Math-7B as the LVLM and LLM foundation models, respectively. Both of the models are reported to have strong basic capabilities.

\noindent \textbf{Training dataset statistics.} GeoMM~\cite{dengRCoTReverseChainThought2024} contains 33344 unique images and 86857 question-answer pairs. Before the supervised fine-tuning process, we embed each question into the corresponding image for augmentation, thus resulting in 86857 pieces of (Image, Question, Description, Solution) data tuple. Geo170k~\cite{gaoGLLaVASolvingGeometric2023} contains 8063 unique images and 117205 question-answer pairs.

\noindent \textbf{Experimental details.}
Our experiments are done with a set of 4 NVIDIA H20 GPUs. The supervised fine-tuning in the first training stage conducts one epoch for both the LVLM and LLM with 128 batch size and $1e-4$ learning rate.
For the reinforcement learning process in the second and third training stages, the number of epochs is 5, the batch size is 64, the learning rate is $1e-6$, and the group number $G = 5$. The hyper-parameters $\varepsilon=0.2$ and the KL coefficient $\beta=0.01$.

\noindent \textbf{Benchmarks.}
We use MMStar~\cite{chen2024we} and MMVet~\cite{yu2023mm} as the benchmark for evaluating the general visual question-answering capabilities of models. MathVista~\cite{luMathVistaEvaluatingMathematical2024} and MathVerse~\cite{zhangMathVerseDoesYour2024} are used as the benchmark for mathematical reasoning. Compared with the previous mathematical benchmarks, MathVerse is a dataset designed to assess the performance of LVLMs in strongly visual-dependent situations. It contains test cases with strong visual dependencies in order to truly verify the LVLM's ability to understand images. As shown in \reffig{verse}, it constructs five different ways of asking for the same question. From the ``Text Dominant'' set to the ``Vision Only'' set, the input becomes more and more visually dependent and requires a higher ability of visual understanding. We use regular expression to extract the final answer from the output response of the model and compare the answer with the ground truth to determine the correctness of the result.

\subsection{Effectiveness of Decoupled Reasoning}
To demonstrate the advantages of the decoupled reasoning method, we compare it with the recent end-to-end vision-language model LLaVA-CoT~\cite{xuLLaVAo1LetVision2024}. With carefully organized data, LLaVA-CoT trains the VLM to independently carry out sequential stages of four sub-tasks: summarization, visual interpretation, logical reasoning, and conclusion generation. To build the decoupled inference pipeline accordingly, we divide the output of the training data of LLaVA-CoT into two parts. The summarization and visual interpretation are used to supervise the dedicated VLM to analyze the question and describe the image. The logical reasoning and conclusion content are used to train the LLM. 

After supervise fine-tuning the VLM for visual interpretation and LLM for linguistic reasoning separately, we evaluate the result. From \reftbl{pre_res}, we can see that the decoupled reasoning method outperforms LLaVA-CoT in general visual question-answering and math vision problems. This shows the rationality of building cross-modal reasoning models based on the advantages of existing mature, specialized models.

\subsection{Effectiveness of Joint-Tuning for Decoupled Reasoning}
\noindent \textbf{Compared methods.} To fully verify the effectiveness of our method, we selected a variety of comparison methods as strong opponents, as shown in \reftbl{main_res}. We compared the current top-performing closed-source models and open-source models and tried two testing strategies: directly using LVLM to answer questions and using LVLM to interpret images and then using LLM to infer and solve questions. Specifically, we use GPT-4o (version ``gpt-4o-2024-11-20'')~\cite{openaiGPT4TechnicalReport2024} as the closed-source LVLM model, GPT-4o \& OpenAI-o1mini (version ``o1-mini-2024-09-12'') as the closed-source LVLM + LLM setting, Qwen2VL-7B as the open-source general LVLM model, R-CoT-7B~\cite{dengRCoTReverseChainThought2024} as the open-source geometry-specific LVLM model, Qwen2VL-72B \& Deepseek-V3(671B) as the open-source LVLM + LLM setting.

For these compared methods, the test setting of lonely LVLM is zero-shot prompting, i.e., we directly instruct the LVLM to think and finally answer the question according to the image. For the LVLM interpretation and LLM reasoning inference pipeline, we prompt LVLM with one in-context example because we found that prompting with an example can demonstrate the image interpretation task to LVLM better and make the output format of LVLM more stable. For the linguistic reasoner LLM, we use zero-shot instruction. For our methods, after we train the LVLM and LLM in three training stages, we use zero-shot prompting for both the LVLM and LLM. The details of the adopted prompts can be found in the supplementary material.

From \reftbl{main_res}, ``GPT-4o \& OpenAI-o1mini'' serves as the closed-source SOTA method surpassing ``GPT-4o'' by a large margin, demonstrating that the decoupled reasoning pipeline bears significant effectiveness in solving geometry math problems. ``Qwen2VL-72B \& Deepseek-V3(671B)'' accordingly demonstrates the effectiveness of open-source models. The open-source geometry-specific LVLM model, R-CoT-7B, has a competitive performance in ``Text Dominant'' and ``Text Lite'' sets of the benchmark. However, its performance drops significantly (e.g., ``Vision Only'' $-23.8$) when vision information becomes harder to extract.

Compared to these opponents, our trained model, which is initialized from ``Qwen2VL-7B \& Qwen2.5Math-7B'', demonstrates advantages, especially in the ``Vision Dominant'' and ``Vision Only'' sets. Equipped with decoupled reasoning architecture, our method can interpret and better understand geometry images with more important conditions related to solving the question. Our method is close to the top-performing open-source method ``Qwen2VL-72B \& Deepseek-V3(671B)'', whose model size is dozens of times larger than ours. The results demonstrate the effectiveness of our decoupled reasoning framework that combines visual interpretation and linguistic reasoning for better geometry problem-solving.

\begin{table}
    \centering
    \caption{Ablation study of image description preprocessing (denoted as $P(\cdot)$).}
    \vspace{-0.6em}
    \label{tab:ab3}
    \setlength{\tabcolsep}{7pt}
    
    \begin{tabular}{l|cc}
    \toprule
        {Test Sets} & {w/o $P(\cdot)$} & {w $P(\cdot)$} \\ \hline
        {Text Dominant} & 47.9 & 52.3 \\ 
        {Text Lite} & 42.6 & 46.9 \\ 
        {Vision Intensive} & 39.5 & 42.2 \\ 
        {Vision Dominant} & 38.8 & 42.6 \\ 
        {Vision Only} & 36.7 & 39.7 \\
    \bottomrule
    \end{tabular}
    \vspace{-0.5em}
\end{table}
\begin{table}
    \centering
    \caption{Ablation study of decoupled reasoning and LVLM only.}
    \vspace{-1em}
    \label{tab:ab1}
    \setlength{\tabcolsep}{7pt}
    
    \begin{tabular}{l|cc}
    \toprule
        {Test Sets} & {LVLM Only} & {Ours} \\ \hline
        {Text Dominant} & 51.1 & 54.3 \\ 
        {Text Lite} & 46.5 & 49.0 \\ 
        {Vision Intensive} & 41.5 & 46.3 \\ 
        {Vision Dominant} & 39.1 & 47.2 \\ 
        {Vision Only} & 37.3 & 43.8 \\
    \bottomrule
    \end{tabular}
    \vspace{-0.5em}
\end{table}

\begin{table}
    \centering
    \caption{Ablation study of the training stages.}
    \vspace{-1em}
    \label{tab:ab2}
    \setlength{\tabcolsep}{7pt}
    
    \begin{tabular}{l|cccc}
    \toprule
        {Stage1(SFT)} & \checkmark & \checkmark & \checkmark & \checkmark \\
        {Stage2(RL-VLM)} & \ding{55} & \checkmark & \ding{55} & \checkmark \\
        {Stage3(RL-LM)} & \ding{55} & \ding{55} & \checkmark & \checkmark \\
        \hline
        {Text Dominant} & 52.3 & 51.5 & 53.2 & 54.3 \\ 
        {Text Lite} & 46.9 & 47.3 & 49.8 & 49.0 \\ 
        {Vision Intensive} & 42.2 & 44.7 & 45.2 & 46.3 \\ 
        {Vision Dominant} & 42.6 & 45.5 & 44.1 & 47.2 \\ 
        {Vision Only} & 39.7 & 42.8 & 40.6 & 43.8 \\
    \bottomrule
    \end{tabular}
    \vspace{-0.5em}
\end{table}

\begin{figure*}[t!]
  \centering
  \includegraphics[width=1\linewidth]{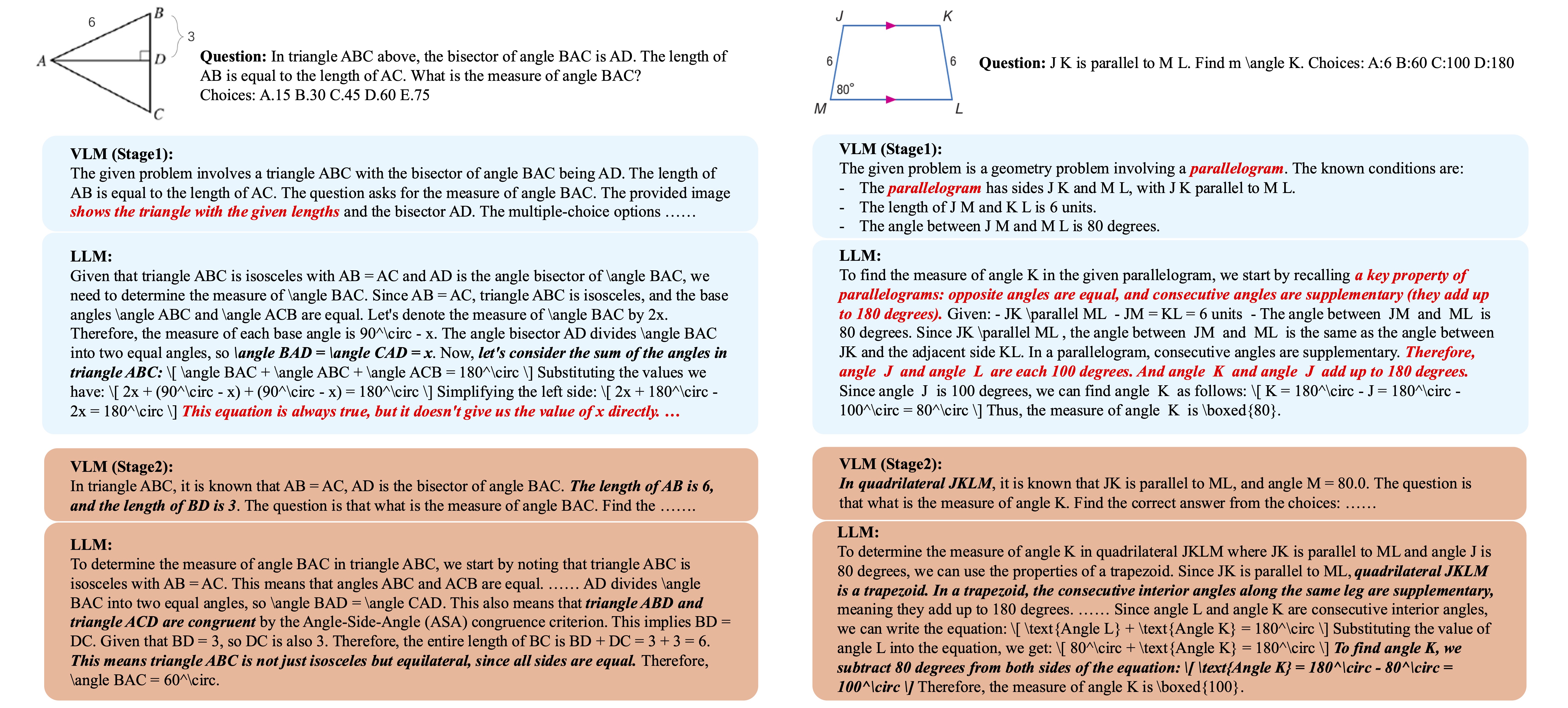}
  \vspace{-2em}
  \caption{The output examples of decoupled vision-language reasoning method. The left example shows that reinforced finetuning can enhance the detail of the generated description (i.e., the exact length of the segment). The right example shows that it can improve the description accuracy (parallelogram v.s. quadrilateral).}
  \label{fig:visualization}
  \vspace{-1em}
\end{figure*}

\begin{figure}[h]
  \centering
  \includegraphics[width=0.98\linewidth]{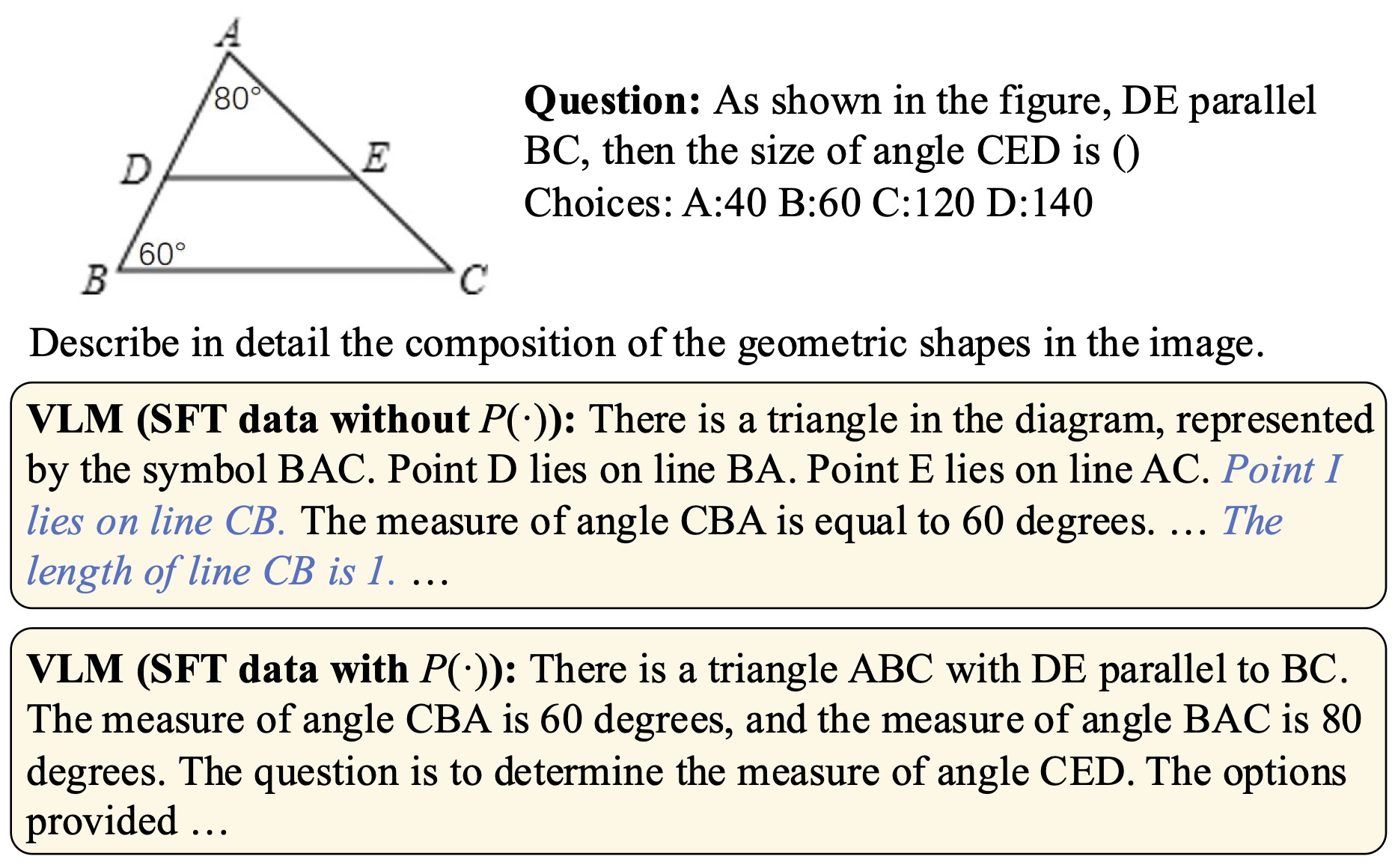}
  \caption{Image interpretation result of the LVLM if preprocessing is not performed.
}
  \label{fig:ab3}
\end{figure}

\subsection{Ablation Study}
\vspace{-0.5em}

To further validate the effectiveness of our method, we conduct an ablation study to examine the contribution of each component. 
We evaluate the performance of the model with the following settings varied:

\noindent \textbf{Effectiveness of image description preprocessing.} 
We ablate the effectiveness of preprocessing the image description data in the SFT stage of VLM. 
The original image descriptions of the GeoMM images are generated by predefined rules, which contain a large number of parallel sentences, like \textit{"Point \{\} lies on line \{\}."}, and \textit{"The length of line \{\} is \{\}."}. This kind of expression, without proper logical order, can easily make the trained VLM generate hallucinatory content with the same sentence structure, as shown in \reffig{ab3}. Besides, the overly verbose and chaotically ordered descriptions may make it more difficult for the LLM to process useful conditions for reasoning.
From \reftbl{ab3}, we can see that preprocessing the original image descriptions from GeoMM has a positive effect on the results of the supervised fine-tuning stage. 

\noindent \textbf{Decoupled Reasoning or LVLM-only.} 
For comparison, we conduct the LVLM-only experiment by supervised fine-tuning using the GeoMM dataset based on the Qwen2VL-7B model. Augmented GeoMM images (according to Section 2.2) are used as input and the reasoning steps of the questions are used as output. After supervised fine-tuning, we then perform reinforcement learning using the Geo170k dataset. During reinforcement learning, the LVLM is tuned by the same outcome-rewarding strategy as Section 2.4. The results in \reftbl{ab1} show the advantages of our decoupled reasoning method over a single LVLM.

Although a single large model that can jointly understand both vision and language input is desirable, we believe that current small-scale LVLMs still have difficulty balancing the exploration capability required for reasoning and the stability required for precise image understanding. Although straightforward, our decoupled reasoning strategy effectively leverages the strengths of explicit visual interpretation and linguistic reasoning, enabling more nuanced understanding and contextual awareness.

\noindent \textbf{Effectiveness of training stages.} 
The results in \reftbl{ab2} examine the effectiveness of each training stage. According to the results, both supervised fine-tuning and reinforcement learning of LVLM on image interpretation lead to clear improvements in the visually dependent test cases. As shown in \reffig{visualization}, supervised fine-tuning enables VLM to interpret images, but it may still miss some details or generate incorrect comprehension. After the second stage of training, VLM can improve the completeness and accuracy of image information extraction.
Performing a complete three-stage joint tuning can further improve and achieve the best results. This confirms the importance of both the visual interpretation and linguistic reasoning modules, as well as the proper collaboration achieved by joint tuning.

\section{Conclusion}
This paper introduces a decoupled vision-language reasoning framework that divides the vision-language problem into visual interpretation and linguistic reasoning.
Visual interpretation specialized VLMs and linguistic reasoning specialized LLMs are optimized to work collaboratively, avoiding end-to-end training of a vision-language model from scratch.
This method presents a cost-efficient solution for multimodal model development. By transforming images into language model-compatible text representations, it facilitates future low-cost and flexible upgrades to upcoming powerful LLMs. Our findings suggest that this framework, preserving the distinct strengths of mature, specialized visual understanding and language reasoning models, forms a valuable baseline for vision-language reasoning. 
Experimental results demonstrate significant improvements in performance on visually intensive geometric math problems, particularly in tasks requiring robust visual understanding. We anticipate that future work will extend this framework to develop VL reasoning capabilities further, enabling a more comprehensive understanding of visual and linguistic information.


\section*{Acknowledgement}
This work was supported by the National Key R\&D Program of China under Grant No. 2023YFA1008500.



{
    \small
    \bibliographystyle{ieeenat_fullname}
    \bibliography{main}
}


\end{document}